\documentclass[sigconf]{acmart}
\usepackage{courier}  
\usepackage{graphicx} 
\usepackage{natbib}  
\usepackage{caption} 

\usepackage{graphicx}
\usepackage{booktabs}
\usepackage{subfigure}
\usepackage{algorithm}
\usepackage{algorithmic}
\usepackage{color}

\usepackage{microtype}
\usepackage{amsmath, amscd, amsfonts}
\usepackage{IEEEtrantools}
\usepackage{soul}
\usepackage{url}
\usepackage{amsmath}
\usepackage{algorithm}
\usepackage{algorithmic}
\usepackage{amsfonts} 
\usepackage{multirow}
\usepackage{boldline}
\usepackage{hhline}
\usepackage{xcolor}

\usepackage{makecell}
\usepackage[inline]{enumitem}
\usepackage{enumitem}
\usepackage{bbding}
\usepackage{pifont}

\AtBeginDocument{%
  \providecommand\BibTeX{{%
    \normalfont B\kern-0.5em{\scshape i\kern-0.25em b}\kern-0.8em\TeX}}}

\copyrightyear{2022}
\acmYear{2022}
\setcopyright{acmcopyright}
\acmConference[SIGIR '22]{Proceedings of the 45th
International ACM SIGIR Conference on Research and Development in Information
Retrieval}{July 11--15, 2022}{Madrid, Spain}
\acmBooktitle{Proceedings of the 45th International ACM SIGIR Conference on
Research and Development in Information Retrieval (SIGIR '22), July 11--15, 2022,
Madrid, Spain}
\acmPrice{15.00}
\acmDOI{10.1145/3477495.3531784}
\acmISBN{978-1-4503-8732-3/22/07}

\begin{document}
\fancyhead{}
\title{Enhancing Event-Level Sentiment Analysis with Structured Arguments}



\author{Qi Zhang}
\affiliation{%
  \institution{Shanghai Key Laboratory of Multidimensional Information Processing, ECNU}
  \country{China}
}
\affiliation{%
  \institution{School of Computer Science and Technology, East China Normal University}
  \city{Shanghai}
  \country{China}
}

\author{Jie Zhou}
\authornote{Jie Zhou is the corresponding authors of this paper.}
\email{jie\_zhou@fudan.edu.cn}
\affiliation{%
  \institution{School of Computer Science, Fudan Univerisity}
  \city{Shanghai}
  \country{China}
 }

\author{Qin Chen}
\affiliation{%
  \institution{School of Computer Science and Technology, East China Normal University}
  \city{Shanghai}
  \country{China}
}

\author{Qinchun Bai}
\affiliation{%
 \institution{Shanghai Engineering Research Center of Open Distance Education, Shanghai Open University}
 \city{Shanghai}
  \country{China}
 }

\author{Liang He}
\affiliation{%
  \institution{School of Computer Science and Technology, East China Normal University}
  \city{Shanghai}
  \country{China}
  }

\renewcommand{\shortauthors}{Q. Zhang and J. Zhou, et al.}

\begin{abstract}
Previous studies about event-level sentiment analysis (SA) usually model the event as a topic, a category or target terms, while the structured arguments (e.g., subject, object, time and location) that have potential effects on the sentiment are not well studied. In this paper, we redefine the task as structured event-level SA and propose an End-to-End Event-level Sentiment Analysis ($\textit{E}^{3}\textit{SA}$) approach to solve this issue. Specifically, we explicitly extract and model the event structure information for enhancing event-level SA. Extensive experiments demonstrate the great advantages of our proposed approach over the state-of-the-art methods. Noting the lack of the dataset, we also release a large-scale real-world dataset with event arguments and sentiment labelling for promoting more researches\footnote{The dataset is available at https://github.com/zhangqi-here/E3SA}.
\end{abstract}

\ccsdesc[500]{Computing methodologies~Natural language processing}
\ccsdesc[100]{Information systems~Web applications}

\keywords{event-level sentiment analysis, structured, datasets}


\maketitle

\section{Introduction}
Sentiment analysis (SA) has received much attention in both academia and industry for its great value \cite{zhou2020sentix,zhou2020modeling}. Instead of assuming that the entire text has an overall sentiment polarity, more and more researchers turn to investigate the fine-grained SA, such as event-level SA \cite{pontiki-etal-2014-semeval,patil2018event,petrescu2019sentiment}.
Event-level SA aims to identify the feelings or opinions expressed by users on a social platform about real-time events from financial news, sports, weather, entertainment, etc. It is vital for many applications, such as stock prediction \cite{makrehchi2013stock}, public opinion analysis \cite{petrescu2019sentiment}. 

\begin{figure}[t!]
    \centering
    \includegraphics[scale=0.14]{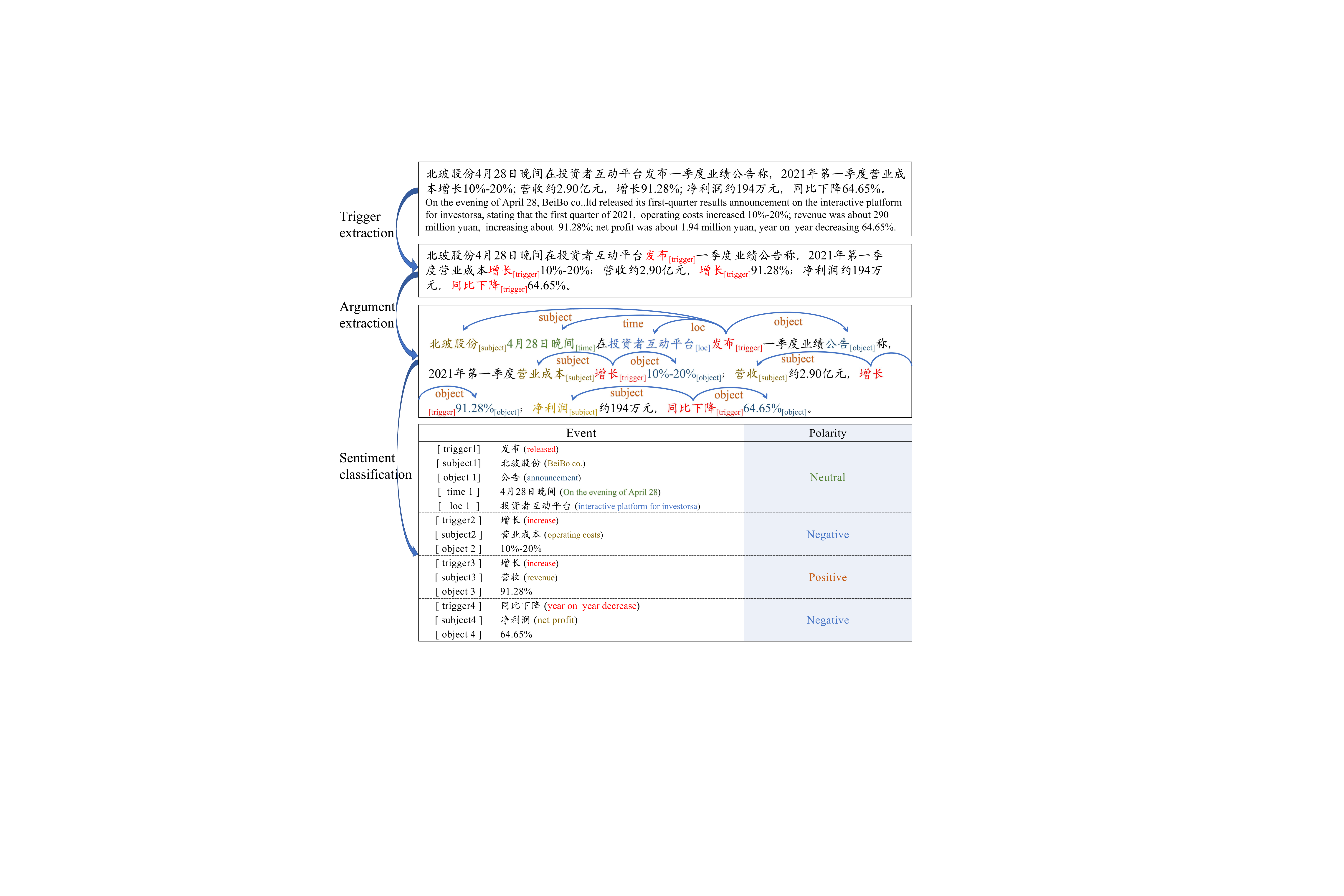}
    \caption{An example of structured event-level SA.}
    \label{fig:example}
\end{figure}

Previous studies about event-level SA mainly utilize the related snippet or context of the event for SA \cite{patil2018event,makrehchi2013stock,fukuhara2007understanding,jagdale2016sentiment,zhou2013sentiment}. \citet{petrescu2019sentiment,patil2018event} detected bursty topics as the events via LDA or clustering algorithm and inferred their sentiments. 
Moreover, \citet{deng2015joint} recognized sentiments toward entities and events that are several terms in the text. Whereas, these studies focused on the event related text modeling, which neglects the influence of the event's inherent structure. 

According to our previous observations, the event structured arguments, such as subject, object, time and location, play an important role in event-level SA. As shown in Figure \ref{fig:example}, the events with the same trigger ``increase" indicate opposite sentiments for different subjects, namely ``operating costs" and ``revenue". Additionally, for the two events (the second and fourth row in the table) with the same negative sentiment, the objects as ``10\%-20\%" and ``64.65\%" help reveal the strength of sentiment polarity. 
Therefore, this work aims to enhance event-level SA with the structured arguments.

Noting that there are few studies about event SA with the fine-grained structured arguments, we reformalize a structured event-level SA task, which extracts the events with arguments and predicts the sentiments. This task suffers from four challenges as follows: 
C1) The multi-subtasks (e.g., trigger extraction, argument extraction, sentiment classification) are related with each other, and performing them independently will lead to error propagation; 
C2) One document may contain multiple events with different sentiments. Taking Figure \ref{fig:example} as an example, there are four events, the sentiment polarities of them are different among each other; 
C3) Unlike general aspect-level SA, the event consists of triggers and arguments associated with their roles, which is harder to model than a topic. Thus, the existing aspect-based SA models can not be applied to this task directly; 
C4) Lack of the labeled datasets for this task. The existing datasets mainly focus on event extraction or aspect's SA, while the sentiment of the structured event is not well studied.

To deal with the above challenges, we present an end-to-end approach for structured event-level SA, which reduces the error propagation among the subtasks (C1). 
Particularly, we first design a feature-enhanced trigger extractor to extract multiple events' triggers simultaneously (C2). 
Second, to better model the event information, we design a trigger-enhanced argument extractor and event-level sentiment classifier, which take trigger and argument information into account (C3).
Finally, we collect and label a real-world dataset in the finance domain for this task (C4). This dataset provides a new perspective for SA and a new benchmark for event-level SA.

The main contributions of this paper are summarized as follows. 
\begin{itemize}[leftmargin=*, align=left]
    \item We reformalize a structured event-level SA task, which focuses on enhancing the event-level SA with structured arguments. 
    \item To mitigate the effect of error propagation, we propose $\textit{E}^3 \textit{SA}$ to model the relationships among the multiple tasks and multiple events by taking structured event elements into account.
    \item For the lack of the off-the-shelf datasets, we release one large-scale dataset for this task. Also, extensive experiments also show that our model outperforms all the strong baselines significantly.
\end{itemize}

\section{Related Work}
In this section, we mainly review the most related papers about event extraction, sentiment analysis and sentiment analysis on events.

\paragraph{\textbf{Event Extraction}}
Event extraction (EE) is a critical task in public opinion monitoring and financial field \cite{li2020unified,yang2019using,DBLP:conf/sigir/LiaoZLZT21,DBLP:conf/sigir/FengLLNC21}. 
It mainly has two key methods, pipeline and joint model.
The pipeline model, which performs trigger extraction and argument role assignment independently \cite{chen2015event,nguyen2015event}, ignoring the relationships among the elements in events.
The joint model generally extracts the related elements at the same time \cite{nguyen2016joint,zhang2019joint,wadden2019entity,li2021document}. 
Recently, researchers have paid attention to document-level EE, which is more complex.
\citet{yang2018dcfee} proposed a DCFEE model to extract the event expressed by multiple sentences based on distant supervision.
\citet{liu2018jointly} and \citet{zheng2019doc2edag} proposed to extract multiple events jointly.
\citet{du-cardie-2020-event} regarded EE as a question answering task to extract the event arguments without entity propagation. Unlike the current work, we focus on modeling the events' sentiment information.

\paragraph{\textbf{Sentiment Analysis}}
Sentiment analysis can be divided into three types: sentence-level, document-level and aspect-level SA \cite{liu2012sentiment,bakshi2016opinion}. 
We mainly review the most related works about aspect-based SA (ABSA) \cite{pontiki-etal-2014-semeval,zhou2019deep}. 
This task aims to predict the sentiments of the aspects in the document, where the aspects are categories, topics or target terms.
To take the aspect into account, attention-based models were designed to capture the relationships between the aspect and its context \cite{fan2018multi,zeng2019lcf}.
Moreover, position information and syntax information was integrated to better model the aspect, such as a category or target \cite{li2018transformation,zhou2020sk}. Applying these models to our task will reduce the performance since we focus on modeling the structured events with complex argument information. 

\paragraph{\textbf{Sentiment Analysis on Events}}
There are some works namely event-based sentiment analysis \cite{patil2018event,makrehchi2013stock,petrescu2019sentiment,fukuhara2007understanding,jagdale2016sentiment,zhou2013sentiment,ebrahimi2017challenges}, which are different from our task. 
Most of these studies focus on detecting the event via topic model and judging the sentiment of the event, which is a category, topic, or term, while the detailed information (e.g., arguments) is ignored by them.
Additionally, they only consider one event in a sentence or document. 
In fact, a text may consist of multiple events and an event is not only a topic. 
To address these problems, we propose an end-to-end approach for event-level sentiment analysis, which aims to identify the events and their sentiments.

\section{Dataset}
\paragraph{\textbf{Data Collection and Annotation}}
Due to lack of annotated resources, we collect and annotate a financial corpus with events and their sentiment polarities, and obtain an event-level SA corpus.
Specifically, we collect 3500 financial news from the portal website\footnote{Note that all the texts are open accessed news on the https://www.eastmoney.com/ without personal information.}. 
We filter the documents that contain less than 50 words or more than 500 words and finally obtain 3500 short documents.
Then, we give annotation guidelines and ten examples to eight human annotators, who manually annotate triggers, arguments and the sentiment labels via Baidu's EasyData platform\footnote{http://ai.baidu.com/easydata/}. 
Note that we only consider the important events that may influence companies' stock or users' decisions.
Because there are too many events in the text and some of the events are not useless for the downstream tasks.
To ensure the labeling quality, each document is annotated by three annotators in order. 
Moreover, we randomly select 100 examples and ask another three annotators to label these documents.
We measure pairwise inter-annotator agreement on tuples among two versions using Krippendorff's alpha coefficient \cite{krippendorff2011computing}.  

\begin{table}[t!]
\centering
\caption{The statistic information of our dataset. 
}
\label{table:dataset}
\scriptsize
\setlength{\tabcolsep}{0.6mm}{\begin{tabular}{lcccccccccc}
\hlineB{4}
      & \#Doc & \#AvgLen & \#E  & \#MultiE & \#PosE & \#NegE & \#NeuE & \#AvgS & \#MultiP & \#E-Across \\  \hline
Train & 2142  & 148.78   & 4210 & 1281   & 2659   & 635    & 916    & 3.41          & 1134           & 474               \\
Dev   & 500   & 151.75   & 962  & 293    & 591    & 154    & 216    & 3.41          & 265            & 104 \\
Test  & 500   & 148.14  & 1005 & 317    & 662    & 138    & 205    & 3.51          & 280            & 122               \\
\hline
Total   & 3142   & 149.15   & 6177  & 1891    & 3912    & 927    & 1337    & 3.42          & 1679            & 593 
\\ 
\hlineB{4}
\end{tabular}}
\end{table}

\begin{table}[t!]
\centering
\scriptsize
\caption{The comparison with the existing datasets. 
}
\label{table:existing datasets}
\setlength{\tabcolsep}{1.0mm}{\begin{tabular}{ll|ccccc}
\hlineB{4}
Task                  & Dataset          & Event & Doc & E-Across & MultiE & Sentiment \\
\hline
\multirow{3}{*}{EE}   & ACE05            &   $\surd$    &  $\surd$        &     -     &    -  &        -   \\
                      & MUC-4 Event      &   $\surd$    &  $\surd$        &      $\surd$      &    $\surd$    &   -        \\
                      & DocEDAG          &   $\surd$    &  $\surd$        &      $\surd$     &    $\surd$    &   -        \\ \hline
\multirow{3}{*}{ABSA} & Twitter          &  -     &    -      &    -      &   -   &    $\surd$       \\
                      & Rest 14 &   -    &     $\surd$     &     -      &   $\surd$     &      $\surd$     \\
                      & Lap 14      &    -   &     $\surd$     &     -    &   $\surd$     &       $\surd$    \\ \hline
Our task                  & Our dataset      &    $\surd$     &     $\surd$     &      $\surd$      &   $\surd$     &       $\surd$    \\   
\hlineB{4}
\end{tabular}}
\end{table}

\paragraph{\textbf{Data Analysis}}
We obtain 3142 samples after filtering the examples without events (Table \ref{table:dataset}).
This dataset has several characteristics: 1) One document always contains multiple events; 2) The events in a document may have different sentiment polarities (\#MultiP); 3) The arguments contained in the same event may be across in different sentences (\#E-Across).
We compare our dataset with the existing ones to clarify the differences. 
First, EE focuses on extracting the events while ignoring their sentiments, and ABSA focuses on the aspects' sentiments while ignoring the structure information. 
Our task aims to judge the events' sentiments, which is more challenging than these two tasks.
Second, though most datasets are at the document level, one event or aspect is always in a sentence. 
In our dataset, one event may be across in multiple sentences. 
Third, our dataset is larger than or comparable with most datasets. ACE2005 and MUC-4 contain less than 1,500 documents.
The size of DOC2EDAG is 32,040, while it is labeled using distant supervision.
\section{Our Approach}
\begin{figure}[t!]
    \centering
    \includegraphics[scale=0.115]{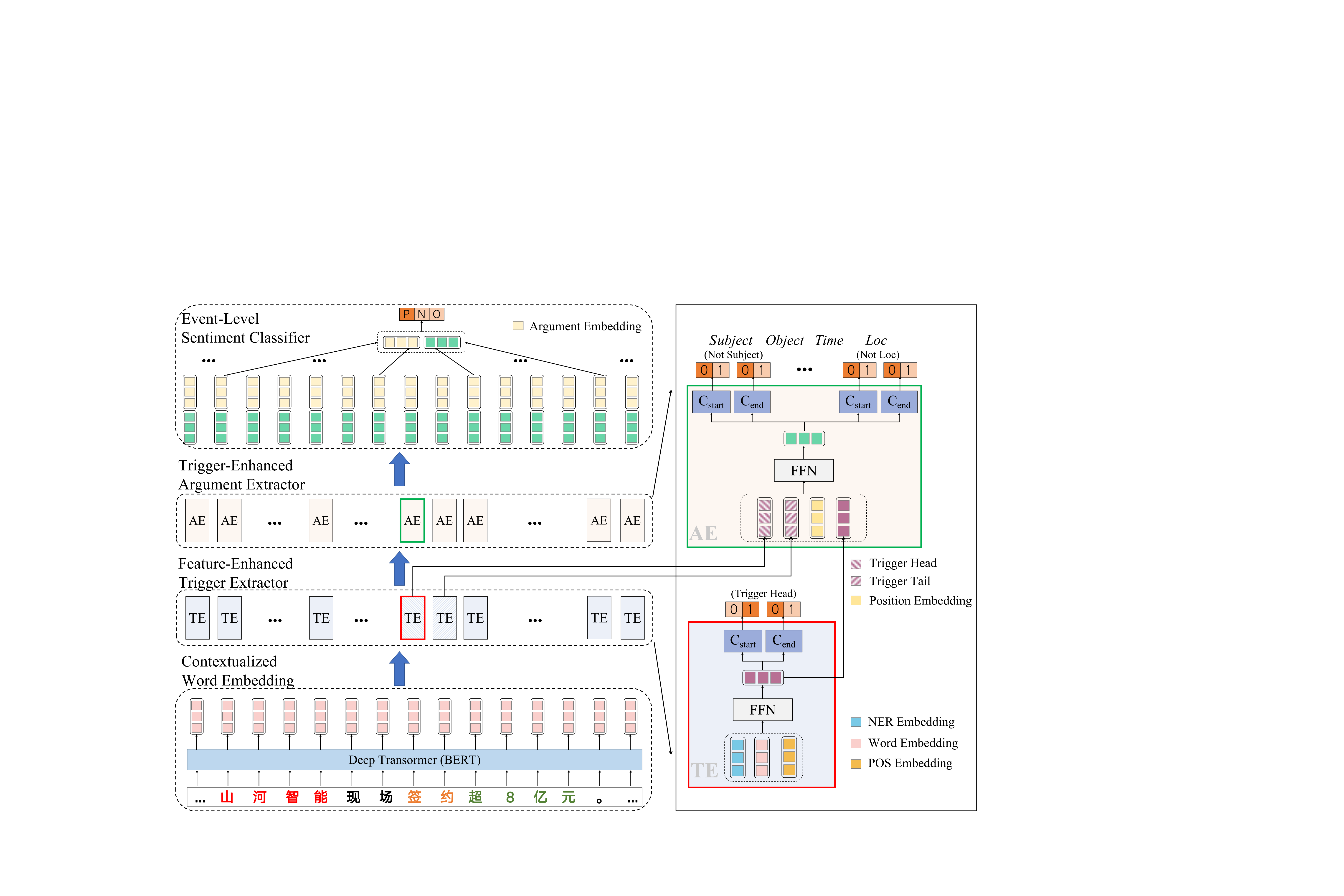}
    \caption{Our $\textit{E}^3 \textit{SA}$ framework.}
    \label{fig:framework}
\end{figure}

In this paper, we propose a $\textit{E}^3 \textit{SA}$ framework for structured event-level SA (Figure \ref{fig:framework}). 
To reduce the propagated errors in the pipeline, 
we propose a joint approach.
$\textit{E}^3 \textit{SA}$ consists of four parts: (i) contextualized word embedding module that models the document with contextual representation; (ii) feature-enhanced trigger extractor that extracts all triggers in the documents with additional features such as POS and NER labels;
(iii) trigger-enhanced argument extractor that extracts the arguments concerning the given trigger by taking trigger information into account; (iv) event-level sentiment classifier that judges the events’ sentiment polarities with argument information.


Formally, given a document $d = \{s_1, ..., s_{|d|}\}$, where $|d|$ is the number of the sentences. $s_i$ is the $i$-th sentence in the document $d$, which contains $|s_i|$ words, $\{w^{(i)}_{1}, ..., w^{(i)}_{|s_i|}\}$. The goal of this task is to extract all the events $E=\{\mathrm{event}_1, ..., \mathrm{event}_{|E|}\}$ in the documents, where the $k$-th event $\mathrm{event}_k=(t_k, \mathrm{a}_k, y_k)$ consists of triggers $t_k$, arguments $a_k$ (subject $\mathrm{sub}_k$, object $\mathrm{obj}_k$, time $\mathrm{time}_k$, location $\mathrm{loc}_k$) and sentiment polarities $y_k$ tuple. The event sentiment polarity $y_k \in \{P, N, O\}$, which represents positive, negative and neutral. 
We aim to maximize the data likelihood of the training set as follows.
\begin{equation}
\nonumber
\scriptsize
\begin{aligned}
    \prod_{k=1}^{|E|}{p(\mathrm{event}_k|d)} = & \prod_{k=1}^{|E|}{p((t_k, \mathrm{a}_k, y_k)|d)} 
     = \prod_{k=1}^{|E|}{p(t_k|d)} p((\mathrm{a}_k, y_k)|t_k, d) \\
      = & \prod_{k=1}^{|E|}{\underbrace{p(t_k|d)}_{\text{Trigger Extraction}}} \underbrace{p(\mathrm{a}_k|t_k, d)}_{\text{Argument Extraction}} \underbrace{p(y_k|t_k, \mathrm{a}_k, d)}_{\text{Sentiment Classification}}
\end{aligned}
\end{equation}

\subsection{Contextualized Word Embedding}
In the word embedding module, we map each word $x_i$ in the input sequence $d$ into a continuous vector space.
Contextualized embedding produced by pre-trained language models \cite{devlin2019bert} have been proved to be capable of improving the performance of a variety of tasks. 
Here, we employ the contextualized representations produced by \textit{BERT-base} to obtain the word embedding.
Specifically, we input the document
$\{\mathrm{[CLS]}, w_{1}, w_{2}, ..., w_{m}, \mathrm{[SEP]}\}$ into \textit{BERT-base}. Then we obtain the word embeddings $\{x^w_{\mathrm{[CLS]}}, x^w_{1}, x^w_{2}, ..., x^w_{m}, x^w_{\mathrm{[SEP]}}\}$, where $m$ is the number of the words in $d$.
where $\mathrm{[CLS]}$ is BERT’s special classification token,
$\mathrm{[SEP]}$ is the special token to denote separation.

\subsection{Feature-Enhanced Trigger Extractor}
Trigger extractor aims to identify whether words trigger an event. We formulate trigger extraction as a token-level classification task and extract all the triggers simultaneously.
We integrate the semantic features (e.g., POS and NER) into text modeling because they are useful for this task. 
For example, most of the triggers are verbs, and most of the arguments are entities and nouns.
Stanza \cite{qi2020stanza} is used to obtain the POS and NER tags of the words.
We forward the concatenation of three types of embedding, including word embedding $x^w$, pos embedding $x^{pos}$ and ner embedding $x^{ner}$ to a feed-forward network (FFN), $x^f_i = \mathrm{FFN}(\mathrm{concat}(x^w_i, x^{pos}_i, x^{ner}_i))$.

Inspired by \cite{wei-etal-2020-novel,yang-etal-2019-exploring}, we train start and end classifiers to enforce the model focus on the triggers' boundaries. The distributions of trigger' start and end are computed as
,
\begin{equation}
\small
\nonumber
\begin{aligned}
    p^{t^s}_{i} = \mathrm{Sigmoid}(W^{t^s}x^f_i+b^{t^s}) ; 
    p^{t^e}_{i} = \mathrm{Sigmoid}(W^{t^e}x^f_i+b^{t^e})
\end{aligned}
\end{equation}
where $^s$ and $^e$ denote the start and end indices, $W^{t^s}$, $W^{t^e}$, $b^{t^s}$ and $b^{t^e}$ are the learnable weights.
As general, we adopt cross entropy (CE) between the predicted probabilities and the ground truth labels as the loss function for fine-tuning.
\begin{equation}
\nonumber
    \mathcal{L}_t = \frac{1}{m}\sum_{i=1}^{m}{\mathrm{CE}(y^{t^s}_i, p^{t^s}_{i})+\mathrm{CE}(y^{t^e}_i, p^{t^e}_{i})}
    \vspace{-1mm}
\end{equation}
where $y^{t^s}_i$/$y^{t^e}_i$ is $1$ if $i$-th word is the trigger's start/end. 

\begin{table*}[t!]
\centering
\caption{The results of event-level SA with extracted arguments. The best scores are marked with bold.}
\label{table:main results}
\scriptsize
\setlength{\tabcolsep}{1.2mm}{\begin{tabular}{l|ccc|ccc|ccc|ccc|ccc|ccc}
\hlineB{4}
          & \multicolumn{3}{c|}{\multirow{2}{*}{Trigger}} & \multicolumn{12}{c|}{Arguments}                                                                         & \multicolumn{3}{c}{\multirow{2}{*}{Sentiment}} \\ 
          & \multicolumn{3}{c|}{}                         & \multicolumn{3}{c}{Sub} & \multicolumn{3}{c}{Obj} & \multicolumn{3}{c}{Time} & \multicolumn{3}{c|}{Loc} & \multicolumn{3}{c}{}                           \\
          & P             & R            & F1            & P      & R     & F1     & P      & R     & F1     & P      & R      & F1     & P      & R     & F1     & P             & R             & F1               \\ \hline
DCFEE-O   &       41.69 & 27.59 & 33.21       & 43.40 & 14.73 & 21.99
&50.79 & 19.30 & 27.97
&71.90 & 48.46 & 57.89
&0.00 & 0.00 & 0.00        &19.75 & 13.07 & 15.73          \\
DCFEE-M     &       33.87& 44.64 & 38.52     &   34.66 & 19.00 & 24.55
&40.81 & 25.17 & 31.14
&58.62 & 59.91 & 59.26
&16.67 & 9.09 & 11.76  &  14.60 & 19.24 & 16.60          \\
GreedyDec  &           \textbf{67.23} & 24.62 & 36.04             &     67.78 & 16.12 & 26.05
&63.74 & 16.62 & 26.36
&79.08 & 53.30 & 63.68
&0.00 & 0.00 & 0.00   &          15.93 & 5.83 & 8.54               \\
Doc2EDAG   &           38.94 & 16.49 & 23.17       &    62.11 & 14.03 & 22.89
&58.75 & 14.03 & 22.65
&56.25 & 11.89 & 19.64
&0.00 & 0.00 & 0.00            &  30.73 & 13.01 & 18.28           \\
BERT-QA   &         51.40 & 60.85 & 55.73      &        69.16 & 55.22 & 61.80
&\textbf{69.96} & 52.84 & 59.20
&75.58 & 57.27 & 65.16
&0.00 & 0.00 & 0.00        &  44.53 & 52.72 & 48.28         \\ \hline
$\textit{E}^3 \textit{SA}$ (Ours)  &           54.79 & \textbf{62.82} & \textbf{58.53}             &       \textbf{69.83} & \textbf{60.80} & \textbf{65.00} & 64.68 & \textbf{55.02} & \textbf{59.46} &\textbf{89.54} & \textbf{60.35} & \textbf{72.11} &\textbf{66.67} & \textbf{18.18} & \textbf{28.57}            &              \textbf{48.24} & \textbf{55.30} & \textbf{51.53}     \\ 
\hlineB{4}        
\end{tabular}}
\end{table*}

\subsection{Trigger-Enhanced Argument Extractor}
Argument extractor aims to identify the related arguments concerning the given trigger.
To better capture the trigger information, we design a trigger-enhanced argument extractor to integrate the trigger's representation and position information into word representation.

For the trigger representation, we use its head and tail word representations.
Also, we define the word position index according to the relative distance with the trigger. 
The word position embedding $x^{\mathrm{position}_k}_{i}$ specific to a trigger $t_k$ can be looked up by a position embedding matrix, which is randomly initialized and updated during the training process. 
Then, we concatenate the trigger's representation and position embedding with feature-enhanced word representation $x^f_i$, and feed them into FFN module, $x^{t_k}_i = \mathrm{FFN}(\mathrm{concat}(x^f_i, x^{f}_{t_k^s}, x^{f}_{t_k^e}, x^{\mathrm{position}_k}_{i}))$,
where $x^{f}_{t_k^s}$ and $x^{f}_{t_k^e}$ is the head and tail representation of the trigger $t_k$ obtained from $x^f$.

Similarly, a word $x_i$ is predicted as the start and end of an argument that plays role $r$ w.r.t. $t_k$ with the probability, 
\begin{equation}
\nonumber
\begin{aligned}
    p^{r_k^s}_{i} = \mathrm{Sigmoid}(W^{r^s}x^{t_k}_i+b^{r^s});
    p^{r_k^e}_{i} = \mathrm{Sigmoid}(W^{r^e}x^{t_k}_i+b^{r^e})
\end{aligned}
\end{equation}

The loss function for argument extraction is, 
\begin{equation}
\nonumber
\scriptsize
\begin{aligned}
     \mathcal{L}_a  = \frac{1}{|E| \times |R| \times m}\sum_{k=1}^{|E|}\sum_{r \in |R|}\sum_{i=1}^{m}{
     {\mathrm{CE}(y^{r_k^s}_i, p^{r_k^s}_{i})+ \mathrm{CE}(y^{r_k^e}_i, p^{r_k^e}_{i})}
     }   
\end{aligned}
\end{equation}
where $R$ is the set of roles, including subject, object, time and location.

\subsection{Event-Level Sentiment Classifier}
Besides the trigger, the arguments information can also help to model the event and its sentiment. 
Thus, we model the argument role into an embedding $x^{r_k}_{i}$, which tells not only the position but also the type information of the arguments. 
We integrate it with trigger-enhanced word representation $x^{t_k}_i$. Then we adopt a max-pooling layer (MaxPooling) to obtain the event representation, $v^{\mathrm{event}_k} = \mathrm{MaxPooling}(\mathrm{concat}(x^{t_k}_i, x^{r_k}_{i}))$.
We input the event representation $v^{\mathrm{event}_k}$ into a softmax layer for sentiment classification, $p_k = \mathrm{Softmax}\\(W^{c}{v^{\mathrm{event}_k}}+b^{c}) $,
where $W^{c}$ and $b^c$ are the learnable parameters.
Then, we calculate the loss of SA, $\mathcal{L}_c = \frac{1}{|E|}\sum_{k=1}^{|E|}\mathrm{CE}{(y_k, p_k)}$.

Finally, to learn the relationships among the multiple subtasks, we learn them jointly by adding the losses together, $\mathcal{L} = \mathcal{L}_t + \mathcal{L}_a + \mathcal{L}_c$.

\section{Experiments}
\subsection{Experimental Setup}
\paragraph{\textbf{Evaluation}}
As for evaluation, we adopt three metrics: precision (P), recall (R), and F1 scores (F1), the same as \cite{li2013joint,zhang2019joint}. 
Additionally, we evaluate the performance of sentiment classification with gold arguments in terms of P, R, F1 and accuracy. 

\paragraph{\textbf{Baselines}}
To verify the effectiveness of our model, we conduct the experiments from two perspectives, end-to-end event extraction and aspect-based SA methods. 
First, we select four widely-used end-to-end event extraction baselines 
to verify their performance on structured event-level SA, including 
DCFEE \cite{yang2018dcfee} (consists of two versions: DCFEE-O and DCFEE-M), GreedyDec \cite{zheng2019doc2edag}, Doc2EDAG \cite{zheng2019doc2edag}, BERT-QA \cite{du-cardie-2020-event}.
For a fair comparison, we replace word embeddings with BERT embeddings for DCFEE. 
Second, we compare our model with four Non-BERT-based and three BERT-based typical ABSA models by inputting the documents and gold events for sentiment prediction, including MemNet \cite{tang2016aspect}, ATAE\_LSTM \cite{wang2016attention}, MGAN \cite{fan2018multi}, TNet \cite{li2018transformation}, BERT-SPC \cite{devlin2019bert}, AEN\_BERT \cite{song2019attentional}, LCF-BERT \cite{zeng2019lcf}.
For the limitation of the space, please see the details of the baselines in the related studies.

\vspace{-2mm}
\paragraph{\textbf{Implementation Details}}
BERT \cite{devlin2019bert} is utilized as the word embedding. 
We use Adam optimizer with the learning rates of 1e-5. The dimensions of position, pos, and ner embedding are 128. The max sequence length is 512. The dropout is 0.1. 
The reported test results are based on the parameters that obtain the best performance on the development set with five random seeds.


\subsection{Main Results}
\paragraph{\textbf{Event-level SA with Extracted Arguments}}
We apply the typical end-to-end event extraction baselines to structured event-level SA task and report the results of these models and $\textit{E}^3 \textit{SA}$ (Table \ref{table:main results}).
From this table, we obtain the following findings. 
\textbf{First}, our model outperforms the strong baselines in most cases. In particular, $\textit{E}^3 \textit{SA}$ obtains better performance than all the baselines significantly in terms of F1 for all the subtasks.
\textbf{Second}, $\textit{E}^3 \textit{SA}$ captures the sentiment information of the events effectively. These baselines focus on predicting the event type while the argument information and the relationships among the events are ignored by these models.
\textbf{Third}, 
most of the models can not extract the location of the event since there are only a few location labels (about 20 times) in the training data. Furthermore, 
our model can extract it more effectively via feature and trigger-enhanced sentence representation. 


\begin{table}[]
\centering
\caption{\small The results of event-level SA with gold arguments.}
\scriptsize
\label{table: event-classification}
\setlength{\tabcolsep}{1.4mm}{\begin{tabular}{ll|cccc}
\hlineB{4}
     &    & P & R & F1 & Acc \\ \hline
\multirow{4}{*}{Non-BERT-based} & MemNet     & 71.25 & 69.65 & 70.41 & 78.41 \\
 & ATAE\_LSTM & 74.84 & 67.92 & 70.72 & 80.00 \\
 & MGAN     &  76.37 &  69.95 &  72.45  &   81.59  \\
 & TNet     &  79.53 &  66.74 &  71.16  &  81.19   \\ \hline
\multirow{3}{*}{BERT-based} & BERT-SPC  &  82.27 & 79.92  & 80.71   &   85.17  \\
 &AEN\_BERT  & 79.94 & 73.11 & 75.93 & 83.18  \\  
 &LCF-BERT  &  81.42 &  80.16 & 80.91   &   85.87  \\ \hline
Ours & $\textit{E}^3 \textit{SA}$    &  \textbf{82.57} &  \textbf{80.24} &  \textbf{81.32}  &  \textbf{86.17}   \\
\hlineB{4}
\end{tabular}}
\end{table}

\paragraph{\textbf{Event-level SA with Gold Arguments}}
To further verify the effectiveness of $\textit{E}^3 \textit{SA}$ on inferring the events' sentiment polarities, we adopt the existing strong baselines of ABSA and perform sentiment classification over structured event-level SA (Table \ref{table: event-classification}). 
We observe that $\textit{E}^3 \textit{SA}$ outperforms all the baselines in terms of F1 and accuracy. 
All the baselines focus on the interaction between the event and the text to capture the event-specific sentiments.
$\textit{E}^3 \textit{SA}$ not only considers the relationships among multiple subtasks but also the relationships among multiple events. 
Moreover, we integrate the trigger and argument information into sentiment classification to capture the sentiment information towards the given events effectively.

\subsection{Ablation Studies}
To further prove the effectiveness of the components contained in $\textit{E}^3 \textit{SA}$, we do ablation studies (Table \ref{table: ablation study}).
First, comparing with the pipeline model (row 2), we find that the end-to-end framework can improve the performance of each subtask by modeling the relationships among the subtasks.
Second, features such as POS and NER can improve the performance effectively because the arguments are always entities and the triggers are always verbs.
Third, removing the trigger information (e.g., trigger head and tail representations, position embedding) will reduce the performance of argument extraction since we aim to extract the argument information w.r.t. the given trigger.
However, the influence of removing the trigger information for sentiment classification is limited because argument information can also help the model learn the event representation.
Fourth, integrating the argument information can capture the sentiment information of the events more effectively.
To further investigate the effectiveness of event information for sentiment classification, we remove both the trigger and argument information from our model, which will reduce the performance significantly.

\begin{table}[]
\centering
\scriptsize
\caption{The results of ablation studies in terms of F1.}
\label{table: ablation study}
\setlength{\tabcolsep}{1.2mm}{\begin{tabular}{l|cccccc}
\hlineB{4}
               & \multirow{2}{*}{Trigger} & \multicolumn{4}{c}{Argument} & \multirow{2}{*}{Sentiment} \\
               &                          & Sub   & Obj   & Time  & Loc  &                            \\ \hline
$\textit{E}^3 \textit{SA}$ (Ours)    &           \textbf{58.53}          &    65.00   &    \textbf{59.46}   &    \textbf{72.11}   &    \textbf{28.57}  &       \textbf{51.53}                     \\
Pipeline       &       56.05                   &    64.89   &  58.16     &   71.22    &    16.67  &         50.25                   \\ \hline
- Feature       &            58.35              &   62.13    &  58.68     &    69.36   &  24.06    &    50.08                        \\
- Trigger Info     &              57.93            &    54.41   &    55.43   &  67.36     &    18.24  &   51.04                         \\
- Argument Info &           58.52               &    \textbf{65.14}   &   58.54    &    71.85   &   27.50   &          50.97                  \\
- Trigger+Argument &           57.20               &    53.07   &   50.06    &    36.43   &   00.00   &          49.58                  \\
\hlineB{4}
\end{tabular}}
\end{table}




\section{Conclusions and Future Work}
In this paper, we propose an effective $\textit{E}^3 \textit{SA}$ approach for structured event-level sentiment analysis.
This joint approach models the relationships among the multi-subtasks and multi-events with structured arguments.
We conduct extensive experiments to evaluate our model on both event extraction and sentiment classification. 
The results demonstrate the great advantages of our model by comparing it with the state-of-the-art baselines.
Additionally, we label a real-world corpus for this task for lack of the off-the-shelf datasets.
It would be interesting to investigate how to integrate users' reviews to better capture the sentiment information of the events.
\begin{acks}
The authors wish to thank the reviewers for
their helpful comments and suggestions. 
This research is funded by the Science and Technology Commission of Shanghai Municipality (No. 19511120200\&2151\\1100100 and 21511100402) and by Shanghai Key Laboratory of Multidimensional Information Processing, East China Normal University, No. 2020KEY001 and the Fundamental Research Funds for the Central Universities. 
This research is also funded by the National Key Research and Development Program of China (No. 2021ZD0114002), the National Nature Science Foundation of China (No. 61906045), and Shanghai Science and Technology Innovation Action Plan International Cooperation project ``Research on international multi language online learning platform and key technologies (No.20510780100)". The computation is performed in ECNU Multi-functional Platform for Innovation (001).
\end{acks}

\bibliographystyle{ACM-Reference-Format}
\bibliography{sample-base}










\end{document}